\documentclass[letterpaper, 10 pt, conference]{ieeeconf}  %

\IEEEoverridecommandlockouts                              %

\usepackage{graphicx} %
\usepackage{times} %
\usepackage{amsmath} %
\usepackage{amssymb}  %
\usepackage{xcolor}
\usepackage{subfig}
\usepackage{multirow}
\usepackage{microtype}
\usepackage{caption}
\usepackage{makecell} %

\makeatletter
\let\NAT@parse\undefined
\makeatother
\usepackage{hyperref}

\newcommand{\vecXX}[1]{{\mathbf {#1}}}

\def \vecq {{\vecXX{q}}}
\def \vecv {{\vecXX{v}}}

\newcommand{\figref}[1]{Figure \ref{fig:#1}}
\newcommand{\tabref}[1]{Table \ref{tab:#1}}
\newcommand{\secref}[1]{$\S$\ref{sec:#1}}

\definecolor{mygreen}{rgb}{0.0, 0.8, 0.0}
\definecolor{myblue}{rgb}{0.4, 0.6, 0.9}
\definecolor{myorange}{rgb}{1.0, 0.71, 0.31}

\title{\LARGE \bf
ReorientBot: Learning Object Reorientation\\for Specific-Posed Placement
}

\author{Kentaro Wada, Stephen James, Andrew J. Davison%
\\
Dyson Robotics Laboratory, Imperial College London%
\\
{\tt\small \{k.wada18, slj12, a.davison\}@imperial.ac.uk}%
}

\begin{document}

\maketitle
\thispagestyle{empty}
\pagestyle{empty}

\begin{abstract}
Robots need the capability of placing objects in arbitrary, specific poses to rearrange the world and achieve various valuable tasks. Object reorientation plays a crucial role in this as objects may not initially be oriented such that the robot can grasp and then immediately place them in a specific goal pose. In this work, we present a vision-based manipulation system, \textit{ReorientBot}, which consists of 1) visual scene understanding with pose estimation and volumetric reconstruction using an onboard RGB-D camera; 2) learned waypoint selection for successful and efficient motion generation for reorientation; 3) traditional motion planning to generate a collision-free trajectory from the selected waypoints. We evaluate our method using the YCB objects in both simulation and the real world, achieving 93\% overall success, 81\% improvement in success rate, and 22\% improvement in execution time compared to a heuristic approach. We demonstrate extended multi-object rearrangement showing the general capability of the system.
\end{abstract}

\section{Introduction}

\begin{figure}[t]
  \centering
  \includegraphics[width=0.97\linewidth]{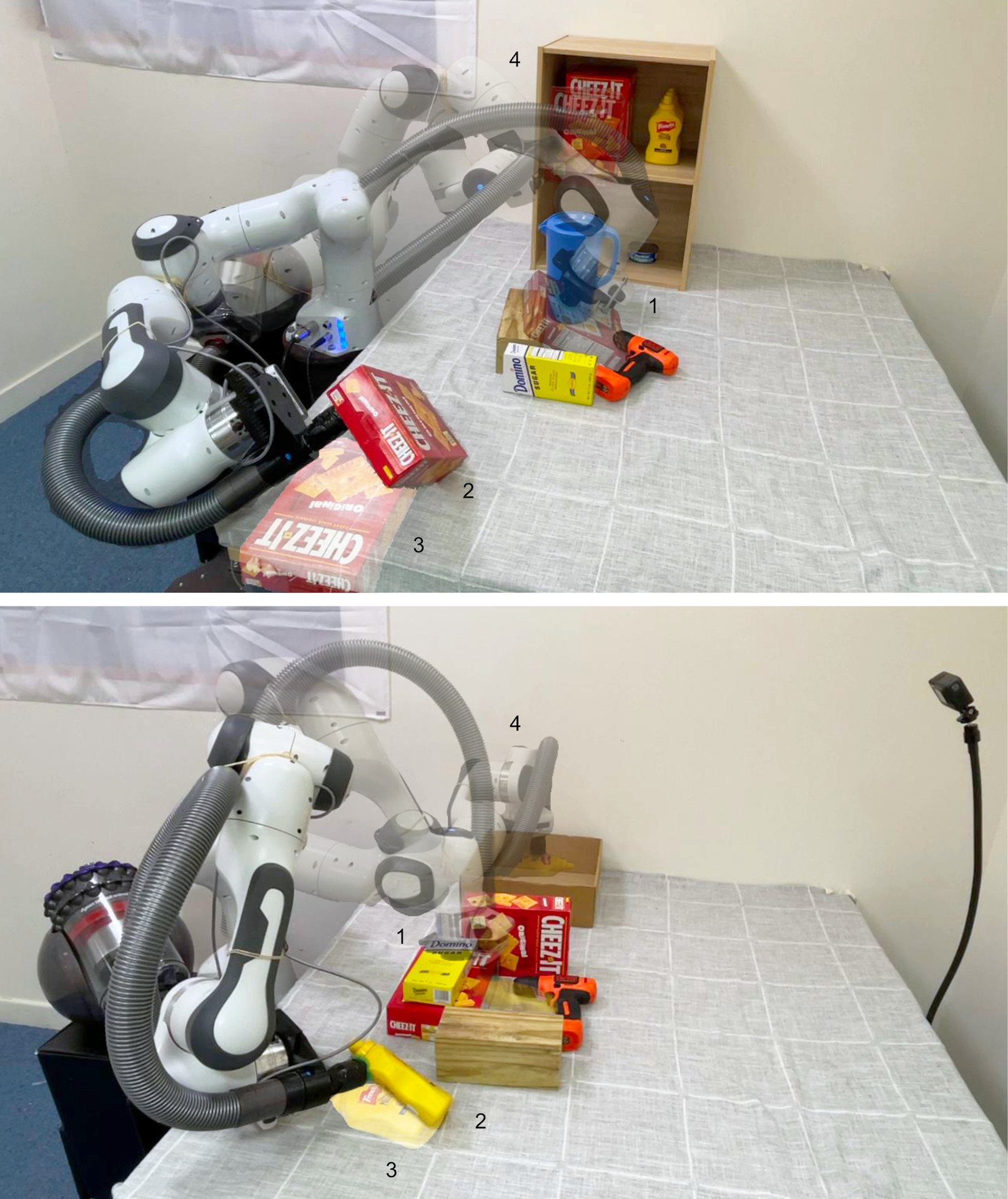}
  \caption{\textbf{ReorientBot} \small{picks, reorients, regrasps, and places objects to rearrange them from a pile to various target configurations. Learned models enable the robot to \textit{dynamically} reorient objects with significant rotation (release and stabilize with gravity), which is hardly achievable with human heuristics.}}
  \label{fig:teaser}
  \vspace{1mm}\hrule\vspace{-6mm}
\end{figure}

Placing objects in a specific pose is a vital capability for robots to rearrange the world to create arbitrary configurations of objects. This capability enables various applications such as product display, storing, or packing, which require tidy, secure, and space-saving object arrangements. When objects must be specifically placed, reorientation is often a crucial manipulation step, to change the object pose in favor of the subsequent steps. Reorientation makes a specific surface of an object accessible when a goal configuration restricts the feasible grasp points, which can be inaccessible in the initial state. With a pile of objects, these grasp points can be blocked by the ground or the surrounding objects, forcing the robot to rotate or flip the object (\figref{teaser}).

Traditionally, object reorientation has been accomplished with hand-designed reorientation poses (e.g., 90-degree rotation), for which a motion planner generates a trajectory~\cite{Tournassoud:etal:ICRA1987, Wan:etal:AR2019}. Although a motion planner can generate a decent trajectory given appropriate and diverse reorientation poses, this approach is often inefficient because of the limited number of pose candidates, requiring multiple reorientation steps for a significant rotation (e.g., flipping). A single-step reorientation would be a solution for this inefficiency; however, it requires careful choices of reorientation poses, which must be both feasible and regraspable. Because of this complex requirement, human heuristics (e.g., canonical, upright reorientation poses) do not achieve high levels of success.

To overcome the limitations of human heuristics for motion generation, an alternative is a learning approach to generate successful and efficient motion trajectories. Although learning approaches for robotic manipulation have become common~\cite{James:etal:ARXIV2021, Levine:etal:JMLR2016} especially in short-horizon tasks such as indiscriminate grasping without precise placement~\cite{Kalashnikov:etal:CORL2018, Levine:etal:IJRR2018}, it is still unclear how to best model long-horizon tasks as it becomes harder to train models as the task horizon increases.

\begin{figure*}[t]
  \centering
  \includegraphics[width=\linewidth]{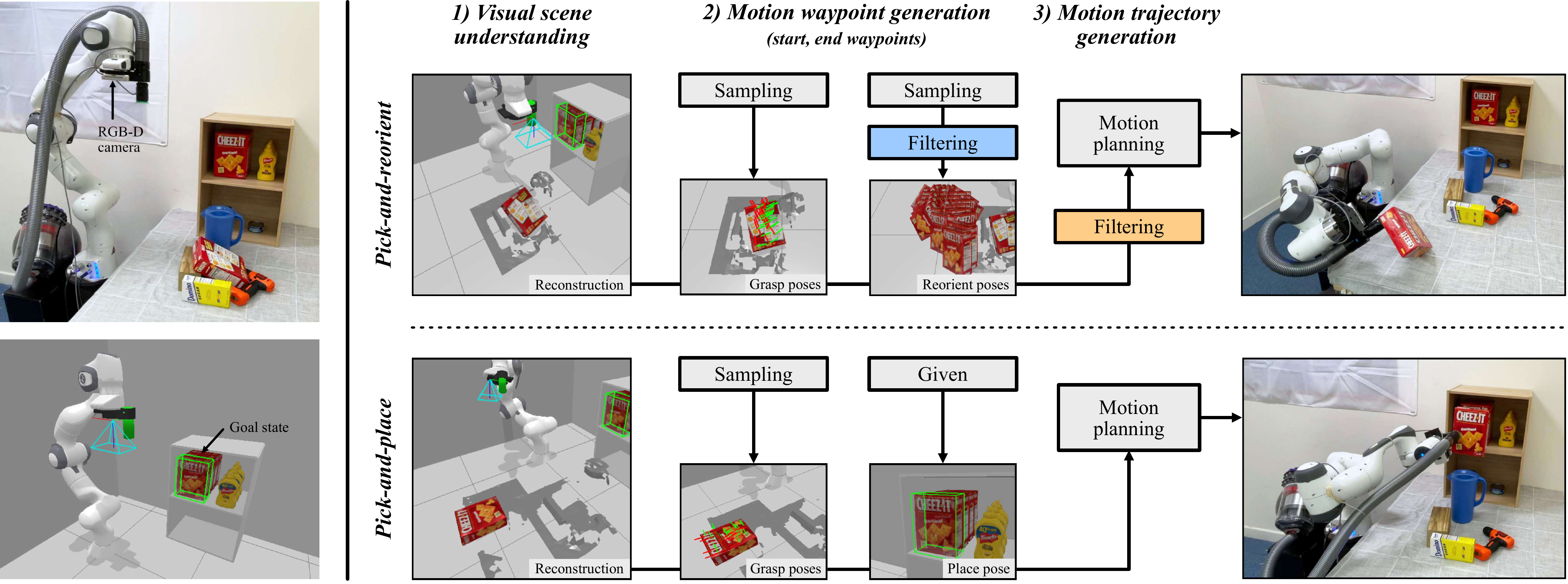}
  \caption{\textbf{System overview}\small{, a hybrid of learned components (\textcolor{myblue}{$\blacksquare$}, \textcolor{myorange}{$\blacksquare$}) and traditional motion planning, consisting of 1) vision-based 6D pose estimation and volumetric reconstruction; 2) motion waypoint generation; 3) trajectory generation using the waypoints.}}
  \label{fig:overview}
  \vspace{1mm}\hrule\vspace{-4mm}
\end{figure*}

Our method uses a sampling-based approach for motion generation, where learned models evaluate the quality of candidate motion waypoints. These learned models predict the success and efficiency of the coarse waypoints, from which trajectories are generated by traditional motion planning. This waypoint evaluation (cf. trajectory evaluation by feeding a long list of waypoints) assumes that the coarse waypoints (e.g., grasp and reorientation pose for object reorientation) stand for the whole trajectory and can be used for evaluation before actually generating the trajectory. This early evaluation drastically reduces the planning time and allows the model to use numerous motion candidates to find the best motion.

Using this approach, we present \textbf{\textit{ReorientBot}}, a hybrid of learned waypoint selection and traditional motion planning with visual scene understanding using a single robot-mounted RGB-D camera (\figref{overview}). The state of a scene is captured by pose estimation for target objects and volumetric reconstruction of non-target objects with a vision system trained for a known object set. This state information is used to generate proposals of the start (grasp/regrasp pose) and end motion waypoints (reorientation/placement pose) with filtering by learned models, which are then fed into motion planning to generate trajectories. The reorientation poses include unstable object states from which the object will settle down to the desired orientation (\textit{dynamic reorientation}), enabling efficient single-step reorientation even for a significant rotation. To our best knowledge, this is the first work that shows dynamic, single-step object reorientation for specific-posed placement with diverse initial and goal states of objects. We demonstrate the capability of the system in the real world showing a real-time scene understanding, planning, and execution.

To summarize, the contributions of this work are:
\begin{itemize}
  \item \textbf{The first work on dynamic, single-step reorientation}, enabling a robot efficiently reorient objects for object rearrangement from an arbitrary initial state to goal state;
  \item \textbf{Learned motion waypoint selection}, taking advantage of the generality of traditional motion planning and the inference speed and robustness of learned models;
  \item \textbf{A full real-time manipulation system}, showing capable object rearrangement with visual scene understanding, learned motion selection, and motion planning.
\end{itemize}

\section{Related work}

\subsection{Robotic pick-and-place}

As a crucial step to isolate objects from a scene, object picking has been widely studied since early robotic research~\cite{Bicchi:Kumar:ICRA2000}. Recent work has integrated vision-based object segmentation and grasp planning to achieve object picking in more challenging, cluttered environments with object overlap and occlusion~\cite{Jonschkowski:etal:IROS2016, Wada:etal:ICRA2022a, Wada:etal:IROS2018, Zeng:etal:ICRA2017}. To handle unseen objects, several studies trained a model to generate object agnostic grasp points~\cite{Pinto:Gupta:ICRA2016, Zeng:etal:ICRA2018} or more general actions such as end-effector transformations~\cite{Kalashnikov:etal:CORL2018, Levine:etal:IJRR2018} from input images. Although these studies showed a strong capability of picking objects in various situations (e.g., occluded, unseen), the placement after picking was mostly simple (e.g., dropping in a box) not knowing how an object is grasped.

A couple of studies have tackled the whole pipeline of robotic pick-and-place, including specific-posed placement. kPAM~\cite{Manuelli:etal:ISRR2019} designed semantic keypoint detection to select a grasp point and plan a placement trajectory, demonstrating intended grasping and specific-posed placement. Shome et al.~\cite{Shome:etal:ICRA2019} showed a tight object packing of box-shaped objects incorporating hand-crafted reorientation motions. These previous studies restricted either object's initial state (e.g., target grasp point is accessible), goal state (e.g., few different orientations), or shape (e.g., box). In this work, we tackle object placement with diverse initial and goal states using various-shaped objects in the YCB object set~\cite{Calli:etal:ICAR2015}.

\subsection{Object reorientation and regrasping}

Robotic research on object reorientation and regrasping dates back to the 1980s with the seminal work by Tournassoud et al.~\cite{Tournassoud:etal:ICRA1987}, and it has been tackled as an essential skill for robotic manipulation~\cite{Cole:etal:TRA1992, Rohrdanz:Wahl:ICRA1997, Wan:Harada:RAL2016}. Several studies demonstrated reorientation and placement via stable object states sampled with object's known~\cite{Lozano:etal:IROS2014, Wan:etal:AR2016, Wan:etal:AR2019} or abstracted geometry (e.g., bounding box)~\cite{Mitash:etal:RAL2020}, or predicted by a learned model~\cite{Cheng:etal:CORL2021, Gualtieri:etal:ICRA2018}. Although they showed successful object reorientation given enough time of execution, they sacrifice the motion efficiency by discarding unstable poses that will eventually become stable in the desired orientation after being released. This restriction makes reorientation with a significant rotation (e.g., flipping) difficult. In this work, we use unstable poses as well as stable poses to plan reorientation to achieve single-step, efficient object reorientation.

In-hand manipulation also has been tackled as a solution to reorient objects to achieve a specific orientation. Dafle et al.~\cite{Dafle:etal:ICRA2014} showed an in-hand regrasping capability with a three-fingered hand such as rolling and flipping. Andrychowicz et al.~\cite{Andrychowicz:etal:IJRR2020} and Akkaya et al.~\cite{Akkaya:etal:ARXIV2019} extended this further to a five-fingered hand to show even more dexterous manipulation such as solving a Rubik's cube. Although promising, the robot's capability heavily depends on the specially designed robotic hand, limiting its applicable environments, object sizes, and poses (e.g., the hand is attached to a fixed base). In this work, we use a suction gripper with a general-purpose robotic manipulator, both of which are widely used for robotic manipulation in industry and research communities.

\section{Method overview}

Given the goal state of target objects, our system runs detection, pose estimation, motion planning to rearrange objects. This system, shown in \figref{overview}, consists of 1) visual 3D scene understanding via 6D pose estimation and volumetric reconstruction; 2) motion waypoint selection that pairs start and end waypoints via learned filtering; 3) trajectory generation by motion planning using the selected waypoints.

This system includes reorientation and regrasping as needed, which is determined via planning the direct pick-and-place from an initial state to a goal state. If this planning fails to find a collision-free motion, the system switches to another motion planner for reorientation. This process is repeated until the motion planner finds a collision-free path for pick-and-place. We optimize the reorientation step to change the object's orientation successfully and efficiently to make the target grasp point accessible to place it in the specified goal state. For this optimization, we sample numerous candidates of reorientation poses, which learned models evaluate via the prediction of feasibility (the existence of a collision-free trajectory) and efficiency (the length of the trajectory).

\section{Visual scene understanding}

Consider a pick-and-place task, with target objects in a pile that a robot must grasp, reorient, regrasp, and place them in a specific pose. This specific-posed placement requires a robot to detect and estimate the target object's initial pose in a pile to compute the relative transformation the robot must apply to achieve the goal state. For non-target objects, semantic scene understanding (detection; pose estimation) might not be as important as for the target since the information is used only for collision avoidance. Therefore, we use a heightmap to represent non-target object's geometry without semantics, allowing faster training and better test-time generalization being agnostic to estimation errors.

\subsection{Object pose for target object}

We run a state-of-the-art object-level mapping system, MoreFusion~\cite{Wada:etal:CVPR2020}, to find target objects using an RGB-D camera mounted on a robotic arm.  MoreFusion consists of learning-based 2D object detection and volumetric 6D pose estimation for the detected objects. Given the class of a target object as a task specification, we retrieve the target's pose (e.g., the initial state in a pile) from the object-level map for the subsequent pipeline.

\subsection{Heightmap for non-target objects}

Given depth images from the RGB-D camera, we build a heightmap, which represents the distance of the object's top surface from the ground at each XY position, to represent the state of non-target objects. We capture the depth images of a pile from the top-down view, and set the pile center to be the center of the heightmap with fixed XY bounds to provide consistent and overall scene information.

\section{Motion waypoint generation}

The start and end configurations of a robot and a target object (i.e., grasp and placement poses) define the waypoints of a motion trajectory. These waypoints are used along with the semantic map of a scene by motion planning to generate a collision-free trajectory. We generate these waypoints by random sampling and learned filtering to select the feasible and efficient motion to execute.

\subsection{Waypoint sampling for pick-and-reorient}

The goal of the \textit{pick-and-reorient} stage is to change the target object orientation such that the robot can grasp specific grasp points to place an object in a specific goal pose (e.g., box packing; shelf storing). A grasp pose represents the start waypoint, and a reorientation pose represents the end waypoint of a reorientation trajectory. We sample the grasp poses on the 3D reconstruction of an object and sample the reorientation poses on an open, planar space near the pile.

\subsubsection{Grasp pose (start waypoint)}\label{sec:pick_and_reorient_grasp_pose}

As the start waypoint for reorientation, grasp poses are sampled on the initial state of a target object in a pile. Given the target object pose from pose estimation, we render the object with a virtual camera in simulation to extract the mask and depth image. We convert the depth image into a point cloud and compute surface normals. Using the object mask aligned to the point cloud and normals, we randomly extract $\sim$30 points and normals on the object surface, which gives the position and quaternion $\vecq = [q_x, q_y, q_z, q_w]^\intercal$ of the grasp pose:
\begin{eqnarray}
  [q_x, q_y, q_z]^\intercal &=& \vecv_g \times \vecv_s \\
  q_w &=& \sqrt{\sum_i{\vecv_{g,i}^2} + \sum_i{\vecv_{s,i}^2}} + (\vecv_g^\intercal \cdot \vecv_s),
\end{eqnarray}
where $\vecv_g$ and $\vecv_s$ are the gripper and surface normal.

\subsubsection{Reorientation pose (end waypoint)}

As the end waypoint, we sample reorientation poses on a planer space adjacent to a pile for efficiency (they could be sampled on any planer surface). Each pose is validated by checking collisions between the CAD model of a target object and the volumetric reconstruction of non-target objects.

Since exhaustive collision checking of arbitrary positions and orientations is time-consuming, we first determine the XY positions where any orientation of the target object will be collision-free. We use a cube with the dimensions of the object's longest axis, which allows an efficient collision checking with the pile reconstruction. To sample the XY positions, we discretize a $\text{0.5m} \times \text{0.3m}$ rectangular space by 10 and 8 each, which provides $10 \times 8 = 80$ candidates. These candidate positions are evaluated with the cube to filter positions that are too close to the pile.

Given the selected XY positions, we compute the Z position and orientation using the actual CAD model of the object instead of the abstracted cube. We discretize the orientation by 8 in each axis of Euler angles, which gives $8^3 = 512$ orientations for each XY position. For each orientation, we compute the distance between the object's bottom and the ground plane and set the Z position to put the object on the plane with a small margin of 2cm. Since we sampled positions where the object's arbitrary orientations would be collision-free, we can reuse the same Z and orientations for other positions. This multistep sampling avoids the slow, combinatorial evaluation of reorientation poses, whose number could be $\text{(XY positions)} \times \text{(orientations)} = 80 \times 512 = 40,960$.

The sampled reorientation poses include unstable states on the plane, which eventually settle down to stable states after being released. These unstable poses allow the robot to reorient objects with a significant rotation in a single step, for example, grasping the backside of an object to flip to the front by leaning the object on the plane while creating a space for the suction gripper as shown in \figref{teaser}.

\subsection{Learning to select reorientation poses}\label{sec:pickable}

\begin{figure}[t]
  \centering
  \includegraphics[width=\linewidth]{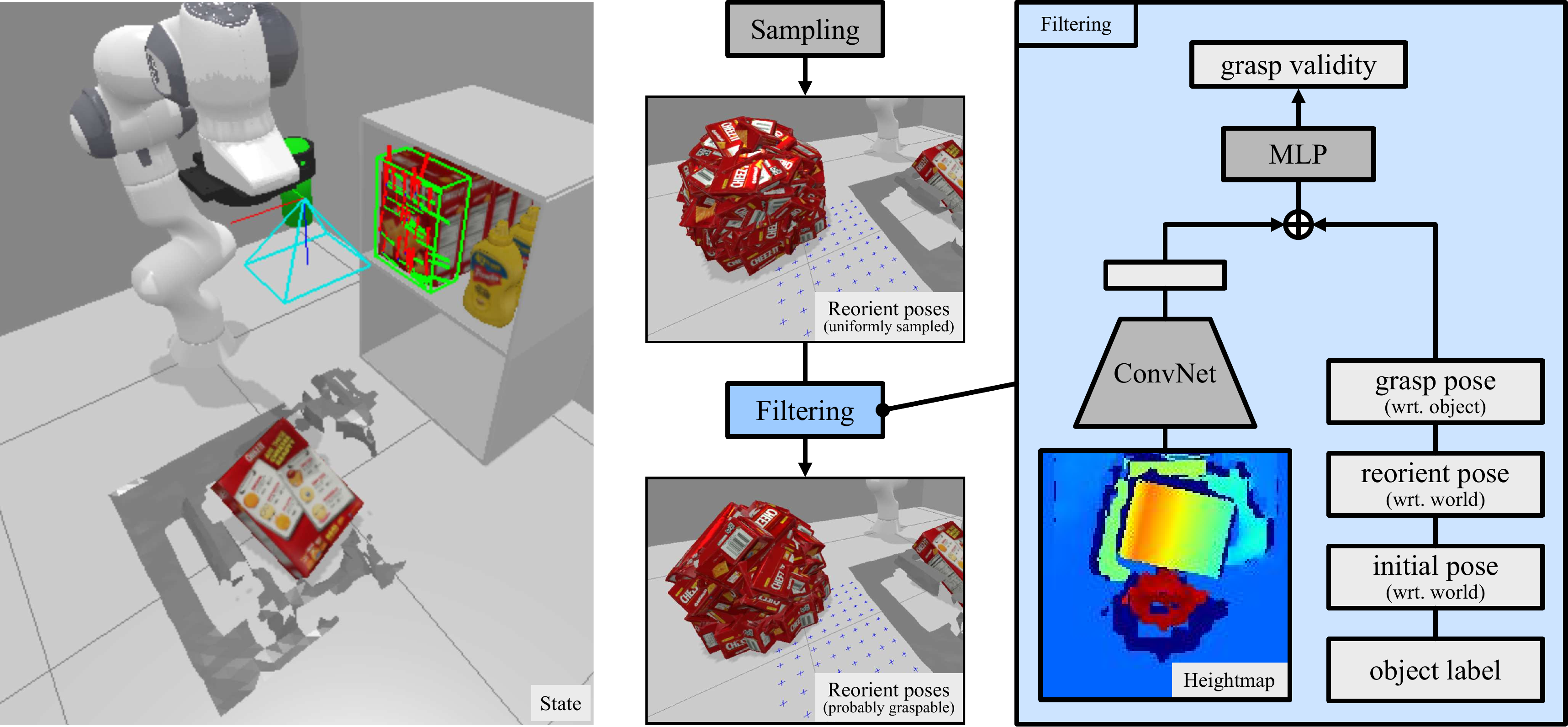}
  \caption{\textbf{Learned filtering of reorientation poses}\small{, which selects feasible (i.e., valid) poses from the uniformly sampled poses.}}
  \label{fig:pickable}
  \vspace{1mm}\hrule\vspace{-5mm}
\end{figure}

Not all the given reorientation poses (40,960 candidates) enable the robot to regrasp the object with the intended grasp for the final placement. To filter these unuseful poses, we introduce a learned model that predicts whether a reorientation pose will enable an intended regrasping as shown in \figref{pickable}. This process uses the model to evaluate reorientation pose candidates and selects the top-1000 best-scored poses to be processed in the subsequent pipeline.

The learned model (right in \figref{pickable}) receives a reorientation and target grasp pose and predicts the success of regrasping after the object is released and settles down. We also feed the pile heightmap to allow the model to take the collisions into account. The model encodes the heightmap with 6 layers of $3\times3$ convolution with max-pooling and ReLU activations (ConvNet). The output is concatenated with the object label, and initial, reorientation, and grasp pose and processed by 3 linear layers (MLP) to predict the grasp validity, trained with binary cross-entropy as a 0--1 probability.

To train the model, we evaluate the reorientation poses using physics simulation and motion planning. With a randomly sampled reorientation pose, the object model is spawned in simulation to apply physics and infer how the object will settle down after being released. Given the stabilized states of reoriented objects, motion planning is applied to test whether the target grasp pose is achievable. This planning result gives the binary label of whether the pair of reorientation pose and grasp pose is feasible (grasp validity in \figref{pickable}), which is used as supervision.

\subsection{Waypoint sampling for pick-and-place}

The goal of the \textit{pick-and-place} stage is to place the target object in the specific pose given as a task goal. The grasp pose represents the start waypoint and the specified final pose represents the end waypoint of the pick-and-place trajectory.

\subsubsection{Grasp pose (start waypoint)}

As the start waypoint for placement, we sample grasp poses from the visible surface of a target object in the goal state with virtual rendering (cf. initial state for reorientation). We position a virtual camera that faces the virtually placed object with a slight translation from the goal state, whose view angle is determined by the direction of the opening of the container: horizontal with shelves, vertical with boxes. By using this virtual rendering, we can sample only the grasp poses visible from the opening of the container while filtering infeasible grasp poses in the back. The grasp position is sampled randomly from the visible surface, and the orientation is determined with the same process as \secref{pick_and_reorient_grasp_pose}, generating $30$ grasp poses as drawn on the goal state of the object in the shelf in \figref{pickable}.

\subsubsection{Placement pose (end waypoint)}

As the end waypoint for placement, we simply use the goal pose of a target object given as a task specification.

\section{Motion trajectory generation}\label{sec:reorientable}

We introduce another learned model to select waypoints for efficient motion planning and exectuion.

\subsection{Learning to select motion waypoints}

\begin{figure}[t]
  \centering
  \includegraphics[width=\linewidth]{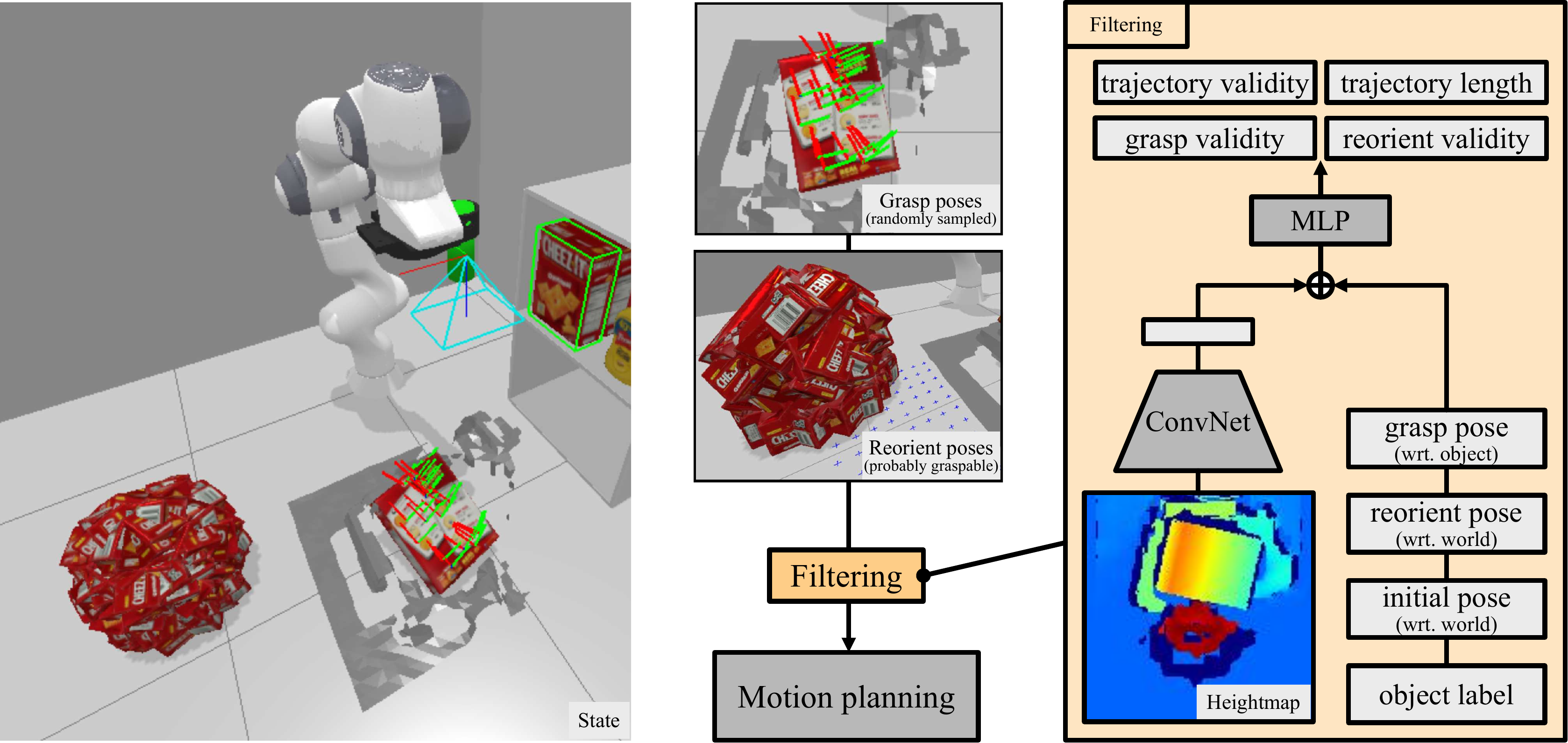}
  \caption{\textbf{Learned filtering of waypoints}\small{, with predicting the validity and efficiency of motion waypoints before feeding them into a motion planner to generate a trajectory.}}
  \label{fig:reorientable}
  \vspace{1mm}\hrule\vspace{-5mm}
\end{figure}

Motion planning runs fast with a few pairs of start and end waypoints (0.1--1.0 seconds) and can generate a collision-free trajectory while evaluating and filtering unusable pairs. However, when the number of pairs becomes large ($>$100), the planning time becomes untenable for real-time use (10--100 seconds). We tackle this problem by introducing a learning-based model that predicts the validity of the waypoint pairs (i.e., the probability that the motion planner finds a collision-free path given those pairs). The low-scored pairs are filtered before feeding them into motion planning (\figref{reorientable}).

With the two motions in the task (pick-and-reorient and pick-and-place), we apply learning-based waypoint selection only to pick-and-reorient. This design choice is because the possible motions for pick-and-place are well constrained by the end waypoint (placement pose), which is unique and given as the task goal (whereas numerous possible reorientation poses must be evaluated for pick-and-reorient), therefore it is not necessary to use a learned model for efficient motion generation. It is also likely that placement configurations vary at test time with different object poses or environments (e.g., shelf storing; box packing), where it would be difficult for a learned model to adapt without retraining.

\subsubsection{Metrics for selection}

The model predicts 3 validities: grasp pose, reorientation pose, and trajectory, each representing the existence of the robot's collision-free state.  Although a single validity could cover the entirety of the grasp and reorientation pose (start and end states) and the trajectory (middle states), we separated these to help the model to reason about why the whole trajectory might be invalid (e.g., which of the start/end/middle states are invalid).

After filtered by validity, several waypoints could remain as candidates with similar predicted validity scores. Therefore, we have introduced another metric: efficiency, which is often regarded as the secondary metric of robotic tasks~\cite{Batra:etal:ARXIV2020}. We use joint-space trajectory length as the efficiency metric, which highly correlates with execution time.

After taking the highest-scored 10 waypoints with the trajectory validity scores, we sort them with efficiency before feeding them into motion planning. Despite the randomness in the planning algorithm while finding collision-free trajectories, we observe a strong correlation between the given waypoints and the generated trajectory (i.e., they are consistent). With this correlation, the learned model predicts meaningful scores to select waypoints that generate the best motion trajectory.

\subsubsection{Model training}

For the waypoint selection, we use a similar model architecture as the reorientation pose selection (right of \figref{reorientable}). Given the start and end waypoints (grasp pose, initial object pose, reorientation pose), this model predicts the validity and efficiency of the trajectory that will be generated by the motion planner, taking the collisions with other objects into account using the heightmap. We train this model with binary cross-entropy loss for the validities and L1 loss for the trajectory length.

\subsection{Collision-free trajectory generation}

Given the selected waypoints, we generate motion trajectories with collision-based motion planning, which uses the scene reconstruction to check feasible states of the robot.

\section{Experiment}

We evaluate our system, ReorientBot, via a set of pick-and-place tasks that require appropriate object reorientation and grasp selection before placing in a specified goal pose. We use 6 large/medium-sized objects (drill, cracker box, sugar box, mustard bottle, pitcher, detergent) in the YCB objects~\cite{Calli:etal:ICAR2015} to evaluate the system in both simulation and the real world.

\subsection{Implementation detail}

We use PyTorch~\cite{Paszke:etal:ANIPS2019} to implement the learned models, training with Adam optimizer~\cite{Kingma:Ba:ICLR2015} with a learning rate of 1e-3.  We stop training as the learning curve converges. For training data collection, we use a physics engine, PyBullet~\cite{Erwin:Yunfei:Misc2016}, to simulate the behavior of objects after being released in unstable reorientation poses for the reorientation pose selection (\secref{pickable}), and stable object pile generation and trajectory evaluation for the waypoints selection (\secref{reorientable}). For the motion planner to generate a collision-free trajectory, we use RRT-Connect~\cite{Kuffner:etal:ICRA2000} implemented with OMPL~\cite{Sucan:etal:2012} integrating with the collision checking on the physics engine.

\subsection{Evaluating in simulation}

\begin{figure}[htbp]
  \vspace{-4mm}
  \centering
  \subfloat[Task configuration]{
    \includegraphics[width=0.403\linewidth]{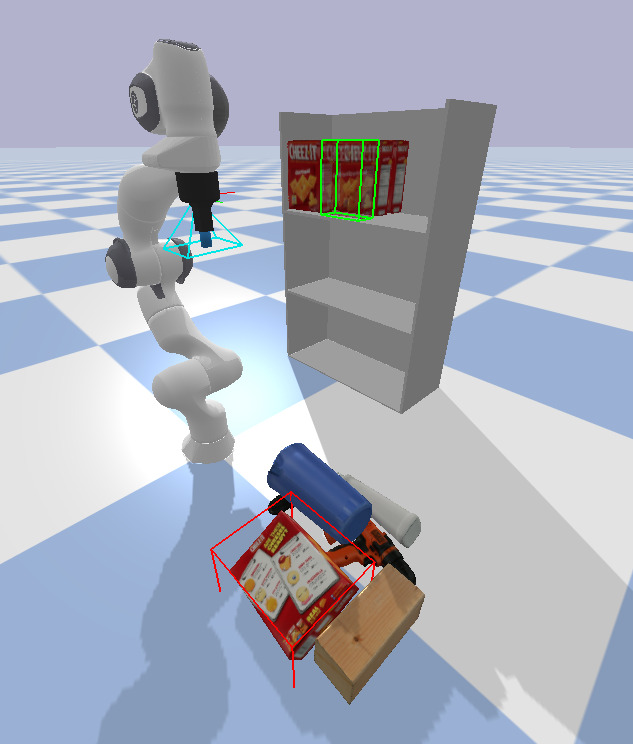}
    \label{fig:evaluation_in_sim_a}
  }
  \subfloat[Heuristic \textcolor{blue}{reorientation poses}]{
    \includegraphics[width=0.53\linewidth]{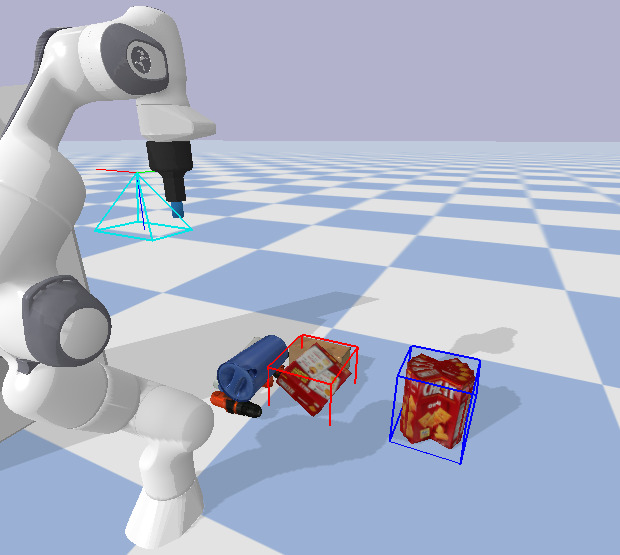}
    \label{fig:evaluation_in_sim_b}
  }
  \caption{\textbf{Evaluation setup in simulation}\small{, tasking the robot to rearrange a target object from an \textcolor{red}{initial state} to a \textcolor{mygreen}{goal state}.}}
  \label{fig:evaluation_in_sim}
  \vspace{1mm}\hrule\vspace{-2mm}
\end{figure}

We evaluate the system in 200 unseen piles. As the goal state, we randomly assign an object pose in the shelf where the same objects are tightly aligned as shown in \figref{evaluation_in_sim}.

We use two types of suction grippers in this experiment:
\begin{itemize}
  \item \textbf{I-shape} (\figref{evaluation_in_sim}), used in previous work~\cite{Zeng:etal:CORL2020}, which has a thin vaccum hose aligned with the gripper;
  \item \textbf{L-shape}, also used in our real-world experiments, where the cup axis is translated from the gripper palm;
\end{itemize}
to show the generality and performance variation of our system. Note that the learned components are trained for each gripper as they are specific to a robot configuration.

As the baseline for single-step object reorientation, we designed heuristic reorientation poses that are stable on a plane and make target grasp points accessible. For simplicity and generality among objects, we use the upright orientation and Z-axis rotation. To make target grasp points accessible after placement, we choose Z-axis orientations where the target grasp points face to -X direction (direction to the robot). \figref{evaluation_in_sim_b} shows the examples of this heuristic reorientation pose for the goal state shown in \figref{evaluation_in_sim_a}.

\begin{figure*}[t]
  \centering
  \includegraphics[width=0.98\linewidth]{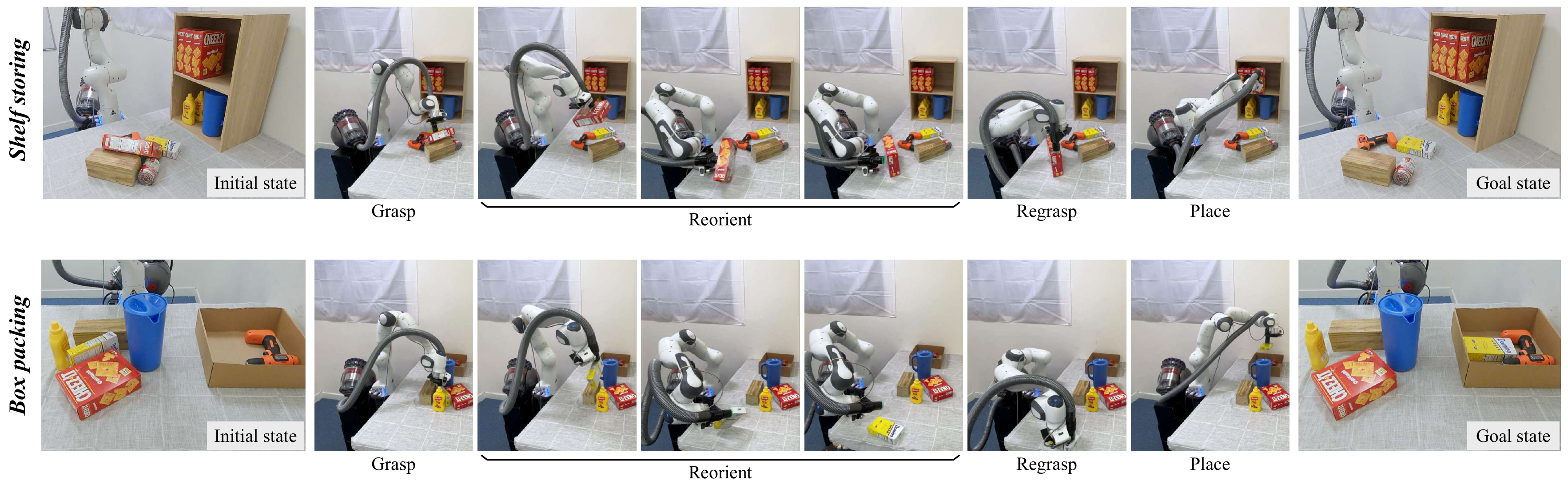}
  \caption{\textbf{Real-world results}\small{, in which the robot rearranges objects from an initial state in a pile to a specified goal state. The reorientation motion includes a dynamic object placement to efficiently accomplish the orientation that enables the final placement.}}
  \label{fig:qualitative}
  \vspace{1mm}\hrule\vspace{-3mm}
\end{figure*}

\subsubsection{Task completion}

\tabref{task_completion} shows the comparison of the success rate, whose criteria is with the geometric distance between the placed state and the goal state, and with 10 seconds time limit for planning. We use the area under the curve (AUC) of the point-to-point distance with a threshold between 0 and 10cm, considering AUC$>$90\% as success (a common metric in pose estimation~\cite{Wada:etal:CVPR2020,Wang:etal:CVPR2019}). The results in \tabref{task_completion} show that, compared to the baseline, ReorientBot gives relative improvements of 23-36\% in reorientation success, 17-47\% in placement success, and 60-81\% in overall success. The success rate with the L-shape gripper is lower, especially in placement because of the more extensive gripper base, which restricts collision-free joint configurations. These improvements in reorientation and placement show that the introduced learned model plays a vital role in both stages of waypoint selection for motion planning.

\begin{table}[htbp]
  \centering
  \caption{\textbf{Task completion}\small{, comparing the baseline (Heuristic) with our method (ReorientBot) in simulation. We use goal configurations that are not achievable without reorientation (146 tasks).}}
  \label{tab:task_completion}
  \begin{tabular}{cc|ccc}
    \textbf{Gripper} & \textbf{Method} & \makecell{\textbf{Success\%}\\\textbf{(reorient)$\uparrow$}} & \makecell{\textbf{Success\%}\\\textbf{(place)$\uparrow$}} & \makecell{\textbf{Success\%}\\\textbf{(overall)$\uparrow$}} \\ \Xhline{2\arrayrulewidth}
    \multirow{2}{*}{I-shape}   & Heuristic & 71.9 & 81.0 & 58.2 \\
                               & ReorientBot & \textbf{97.9} & \textbf{95.1} & \textbf{93.2} \\ \hline
    \multirow{2}{*}{L-shape} & Heuristic   & 74.0 & 58.3 & 43.2 \\
                             & ReorientBot & \textbf{91.1} & \textbf{85.7} & \textbf{78.1} \\
  \end{tabular}
  \vspace{-1mm}
\end{table}

\subsubsection{Timing}

\tabref{task_timing} shows the comparison of the planning and execution time in reorientation. We measure the planning time with the wall clock and execution time with the simulation clock. We send the motion trajectory (a list of joint positions) to the position controller with a constant speed of $\sim$1.2 rad/s (=70 deg/s). As this execution speed can vary in the real world, we also report the trajectory length, which is highly correlated. ReorientBot gives improvements of 24-30\% in planning time and 20-22\% in execution time compared to the baseline. This result shows that the learned filters allow the motion planner to evaluate only the promising waypoints, which are likely to provide a valid and efficient trajectory.

\begin{table}[htbp]
  \centering
  \caption{\textbf{Timing}\small{, comparing the baseline (Heuristic) with our method (ReorientBot) in simulation. Reporting only when both methods succeeded to complete task; 38 tasks out of 146 in \tabref{task_completion}.}}
  \label{tab:task_timing}
  \begin{tabular}{cc|ccc}
    \textbf{Gripper} & \textbf{Method} & \makecell{\textbf{Planning}\\\textbf{time [s]$\downarrow$}} & \makecell{\textbf{Execution}\\\textbf{time [s]$\downarrow$}} & \makecell{\textbf{Trajectory}\\\textbf{length [rad]$\downarrow$}} \\ \Xhline{2\arrayrulewidth}
    \multirow{2}{*}{I-shape} & Heuristic & 3.3 & 4.0 & 4.2 \\
                             & ReorientBot & \textbf{2.5} & \textbf{3.2} & \textbf{3.3} \\ \hline
    \multirow{2}{*}{L-shape} & Heuristic & 3.0 & 3.6 & 3.6\\
                             & ReorientBot & \textbf{2.0} & \textbf{2.8} & \textbf{2.7} \\
  \end{tabular}
  \vspace{-4mm}
\end{table}

\subsection{Real-world evaluation}

We evaluate our system in the real world integrating a Franka Emika Panda robot with the Robotic Operation System framework~\cite{Quigley:etal:ICRA2009}. To capture short-range depths ($\sim$0.1m), we use a Realsense D435~\cite{Keselman:etal:CVPRW2017} as the onboard RGB-D camera. Qualitative results are best seen via supplementary videos.

\figref{qualitative} shows the sequences of the pick and place motions of the robot for the specified goal configuration in a shelf and box. These examples show the capability of our system to reorient objects both successfully and efficiently (with a short arm trajectory), including a dynamic reorientation of objects. They also demonstrate precise placement (e.g., inserting the yellow box into the narrow gap of the drill) and generality in various goal configurations (side, top-down placement).

\section{Conclusion}

We have presented a robotic system that can rearrange objects to a specific goal state, including reorientation and regrasping for final placement. Our system integrates learned waypoint selection and traditional motion planning to maintain capability by learning (efficient selection of motion waypoints) and generality by planning (flexible motion generation based on a goal state and constraints at test time). The resulting system improves on a baseline in both efficiency and success rate, and has shown capable, dynamic reorientation for significant rotation (e.g., flipping) and precise placement in various target configurations (shelf storing; box packing).

We believe there are still various possibilities in combining learning models with traditional motion planning. In this work, the learned models evaluate only two waypoints (start, end) to optimize the reorientation motion. However, the optimization of longer-horizon tasks such as a whole rearrangement pipeline (grasp, reorient, regrasp, place) and tasks with navigation and multiple robots would require a more fine-grained indication by learning to the motion planner.

\section*{Acknowledgements}
Research presented in this paper has been supported by Dyson Technology Ltd.

\bibliographystyle{plain}
\bibliography{robotvisiontex/robotvision}

\begin{thebibliography}{10}

\bibitem{Akkaya:etal:ARXIV2019}
Ilge Akkaya, Marcin Andrychowicz, Maciek Chociej, Mateusz Litwin, Bob McGrew,
  Arthur Petron, Alex Paino, Matthias Plappert, Glenn Powell, Raphael Ribas,
  et~al.
\newblock Solving rubik's cube with a robot hand.
\newblock {\em arXiv preprint arXiv:1910.07113}, 2019.

\bibitem{Andrychowicz:etal:IJRR2020}
Marcin Andrychowicz, Bowen Baker, Maciek Chociej, Rafal Jozefowicz, Bob McGrew,
  Jakub Pachocki, Arthur Petron, Matthias Plappert, Glenn Powell, Alex Ray,
  et~al.
\newblock Learning dexterous in-hand manipulation.
\newblock {\em {International Journal of Robotics Research ({IJRR})}},
  39(1):3--20, 2020.

\bibitem{Batra:etal:ARXIV2020}
D.~Batra, A.~X. Chang, S.~Chernova, A.~J. Davison, J.~Deng, V.~Koltun,
  S.~Levine, J.~Malik, I.~Mordatch, R.~Mottaghi, M.~Savva, and H.~Su.
\newblock Rearrangement: A challenge for embodied {AI}.
\newblock {\em arXiv preprint arXiv:2011.01975}, 2020.

\bibitem{Bicchi:Kumar:ICRA2000}
Antonio Bicchi and Vijay Kumar.
\newblock Robotic grasping and contact: A review.
\newblock In {\em {Proceedings of the {IEEE} International Conference on
  Robotics and Automation ({ICRA})}}, 2000.

\bibitem{Calli:etal:ICAR2015}
B.~Calli, A.~Singh, A.~Walsman, P~Srinivasa S.~and, Abbeel, and A.~M. Dollar.
\newblock {The YCB object and Model set}: Towards common benchmarks for
  manipulation research.
\newblock In {\em International Conference on Advanced Robotics (ICAR)}, 2015.

\bibitem{Cheng:etal:CORL2021}
Shuo Cheng, Kaichun Mo, and Lin Shao.
\newblock Learning to regrasp by learning to place.
\newblock In {\em {Conference on Robot Learning ({CoRL})}}, 2021.

\bibitem{Cole:etal:TRA1992}
Arlene~A Cole, Ping Hsu, and S~Shankar Sastry.
\newblock Dynamic control of sliding by robot hands for regrasping.
\newblock {\em {{IEEE} Transactions on Robotics and Automation}}, 1992.

\bibitem{Erwin:Yunfei:Misc2016}
Erwin Coumans and Yunfei Bai.
\newblock {PyBullet}, a python module for physics simulation for games,
  robotics and machine learning.
\newblock \url{http://pybullet.org}, 2016--2021.

\bibitem{Dafle:etal:ICRA2014}
Nikhil~Chavan Dafle, Alberto Rodriguez, Robert Paolini, Bowei Tang,
  Siddhartha~S Srinivasa, Michael Erdmann, Matthew~T Mason, Ivan Lundberg,
  Harald Staab, and Thomas Fuhlbrigge.
\newblock Extrinsic dexterity: In-hand manipulation with external forces.
\newblock In {\em {Proceedings of the {IEEE} International Conference on
  Robotics and Automation ({ICRA})}}, 2014.

\bibitem{Gualtieri:etal:ICRA2018}
Marcus Gualtieri, Andreas ten Pas, and Robert Platt.
\newblock Pick and place without geometric object models.
\newblock In {\em {Proceedings of the {IEEE} International Conference on
  Robotics and Automation ({ICRA})}}, 2018.

\bibitem{James:etal:ARXIV2021}
Stephen James, Kentaro Wada, Tristan Laidlow, and Andrew~J Davison.
\newblock {Coarse-to-Fine Q-attention}: Efficient learning for visual robotic
  manipulation via discretisation.
\newblock {\em arXiv preprint arXiv:2106.12534}, 2021.

\bibitem{Jonschkowski:etal:IROS2016}
Rico Jonschkowski, Clemens Eppner, Sebastian H{\"o}fer, Roberto
  Mart{\'\i}n-Mart{\'\i}n, and Oliver Brock.
\newblock Probabilistic multi-class segmentation for the amazon picking
  challenge.
\newblock In {\em {Proceedings of the {IEEE/RSJ} Conference on Intelligent
  Robots and Systems ({IROS})}}, 2016.

\bibitem{Kalashnikov:etal:CORL2018}
Dmitry Kalashnikov, Alex Irpan, Peter Pastor, Julian Ibarz, Alexander Herzog,
  Eric Jang, Deirdre Quillen, Ethan Holly, Mrinal Kalakrishnan, Vincent
  Vanhoucke, et~al.
\newblock {QT-Opt}: Scalable deep reinforcement learning for vision-based
  robotic manipulation.
\newblock In {\em {Conference on Robot Learning ({CoRL})}}, 2018.

\bibitem{Keselman:etal:CVPRW2017}
Leonid Keselman, John Iselin~Woodfill, Anders Grunnet-Jepsen, and Achintya
  Bhowmik.
\newblock Intel realsense stereoscopic depth cameras.
\newblock In {\em {Proceedings of the {IEEE} Conference on Computer Vision and
  Pattern Recognition Workshops ({CVPRW})}}, 2017.

\bibitem{Kingma:Ba:ICLR2015}
Diederik~P. Kingma and Jimmy Ba.
\newblock Adam: {A} method for stochastic optimization.
\newblock In {\em {Proceedings of the International Conference on Learning
  Representations ({ICLR})}}, 2015.

\bibitem{Kuffner:etal:ICRA2000}
James~J Kuffner and Steven~M LaValle.
\newblock {RRT-connect}: An efficient approach to single-query path planning.
\newblock In {\em {Proceedings of the {IEEE} International Conference on
  Robotics and Automation ({ICRA})}}, 2000.

\bibitem{Levine:etal:JMLR2016}
Sergey Levine, Chelsea Finn, Trevor Darrell, and Pieter Abbeel.
\newblock End-to-end training of deep visuomotor policies.
\newblock {\em The Journal of Machine Learning Research}, 17(1), Jan 2016.

\bibitem{Levine:etal:IJRR2018}
Sergey Levine, Peter Pastor, Alex Krizhevsky, Julian Ibarz, and Deirdre
  Quillen.
\newblock Learning hand-eye coordination for robotic grasping with deep
  learning and large-scale data collection.
\newblock {\em {International Journal of Robotics Research ({IJRR})}},
  37(4-5):421--436, 2018.

\bibitem{Lozano:etal:IROS2014}
Tom{\'a}s Lozano-P{\'e}rez and Leslie~Pack Kaelbling.
\newblock A constraint-based method for solving sequential manipulation
  planning problems.
\newblock In {\em {Proceedings of the {IEEE/RSJ} Conference on Intelligent
  Robots and Systems ({IROS})}}, 2014.

\bibitem{Manuelli:etal:ISRR2019}
Lucas Manuelli, Wei Gao, Peter Florence, and Russ Tedrake.
\newblock {kPAM}: Keypoint affordances for category-level robotic manipulation.
\newblock {\em {Proceedings of the International Symposium on Robotics Research
  ({ISRR})}}, 2019.

\bibitem{Mitash:etal:RAL2020}
Chaitanya Mitash, Rahul Shome, Bowen Wen, Abdeslam Boularias, and Kostas
  Bekris.
\newblock Task-driven perception and manipulation for constrained placement of
  unknown objects.
\newblock {\em {{IEEE} Robotics and Automation Letters}}, 5(4):5605--5612,
  2020.

\bibitem{Paszke:etal:ANIPS2019}
Adam Paszke, Sam Gross, Francisco Massa, Adam Lerer, James Bradbury, Gregory
  Chanan, Trevor Killeen, Zeming Lin, Natalia Gimelshein, Luca Antiga, et~al.
\newblock {PyTorch}: An imperative style, high-performance deep learning
  library.
\newblock In {\em Advances in neural information processing systems}, 2019.

\bibitem{Pinto:Gupta:ICRA2016}
Lerrel Pinto and Abhinav Gupta.
\newblock Supersizing self-supervision: Learning to grasp from 50k tries and
  700 robot hours.
\newblock In {\em {Proceedings of the {IEEE} International Conference on
  Robotics and Automation ({ICRA})}}, 2016.

\bibitem{Quigley:etal:ICRA2009}
Morgan Quigley, Ken Conley, Brian Gerkey, Josh Faust, Tully Foote, Jeremy
  Leibs, Rob Wheeler, and Andrew~Y Ng.
\newblock {ROS}: an open-source robot operating system.
\newblock In {\em In Proceedings of the {IEEE} International Conference on
  Robotics and Automation Workshops (ICRAW)}, 2009.

\bibitem{Rohrdanz:Wahl:ICRA1997}
F~Rohrdanz and Friedrich~M Wahl.
\newblock Generating and evaluating regrasp operations.
\newblock In {\em {Proceedings of the {IEEE} International Conference on
  Robotics and Automation ({ICRA})}}, 1997.

\bibitem{Shome:etal:ICRA2019}
Rahul Shome, Wei~N Tang, Changkyu Song, Chaitanya Mitash, Hristiyan Kourtev,
  Jingjin Yu, Abdeslam Boularias, and Kostas~E Bekris.
\newblock Towards robust product packing with a minimalistic end-effector.
\newblock In {\em {Proceedings of the {IEEE} International Conference on
  Robotics and Automation ({ICRA})}}, 2019.

\bibitem{Sucan:etal:2012}
Ioan~Alexandru Sucan, Mark Moll, and Lydia~E Kavraki.
\newblock The open motion planning library.
\newblock {\em Robotics \& Automation Magazine, IEEE}, 19(4):72--82, 2012.

\bibitem{Tournassoud:etal:ICRA1987}
Pierre Tournassoud, Tom{\'a}s Lozano-P{\'e}rez, and Emmanuel Mazer.
\newblock Regrasping.
\newblock In {\em {Proceedings of the {IEEE} International Conference on
  Robotics and Automation ({ICRA})}}, 1987.

\bibitem{Wada:etal:ICRA2022a}
Kentaro Wada, Stephen James, and Andrew~J. Davison.
\newblock {SafePicking}: Learning safe object extraction via object-level
  mapping.
\newblock In {\em {Proceedings of the {IEEE} International Conference on
  Robotics and Automation ({ICRA})}}, 2022.

\bibitem{Wada:etal:IROS2018}
Kentaro Wada, Shingo Kitagawa, Kei Okada, and Masayuki Inaba.
\newblock Instance segmentation of visible and occluded regions for finding and
  picking target from a pile of objects.
\newblock In {\em {Proceedings of the {IEEE/RSJ} Conference on Intelligent
  Robots and Systems ({IROS})}}, 2018.

\bibitem{Wada:etal:CVPR2020}
Kentaro Wada, Edgar Sucar, Stephen James, Daniel Lenton, and Andrew~J. Davison.
\newblock {MoreFusion}: Multi-object reasoning for {6D} pose estimation from
  volumetric fusion.
\newblock In {\em {Proceedings of the {IEEE} Conference on Computer Vision and
  Pattern Recognition ({CVPR})}}, 2020.

\bibitem{Wan:etal:AR2016}
Weiwei Wan and Kensuke Harada.
\newblock Achieving high success rate in dual-arm handover using large number
  of candidate grasps, handover heuristics, and hierarchical search.
\newblock {\em Advanced Robotics}, 30(17-18):1111--1125, 2016.

\bibitem{Wan:Harada:RAL2016}
Weiwei Wan and Kensuke Harada.
\newblock Developing and comparing single-arm and dual-arm regrasp.
\newblock {\em {{IEEE} Robotics and Automation Letters}}, 1:243--250, 2016.

\bibitem{Wan:etal:AR2019}
Weiwei Wan, Hisashi Igawa, Kensuke Harada, Hiromu Onda, Kazuyuki Nagata, and
  Natsuki Yamanobe.
\newblock A regrasp planning component for object reorientation.
\newblock {\em Autonomous Robots}, 43(5):1101--1115, 2019.

\bibitem{Wang:etal:CVPR2019}
Chen Wang, Danfei Xu, Yuke Zhu, Roberto Mart{\'\i}n-Mart{\'\i}n, Cewu Lu,
  Li~Fei-Fei, and Silvio Savarese.
\newblock {DenseFusion}: {6D} object pose estimation by iterative dense fusion.
\newblock {\em {Proceedings of the {IEEE} Conference on Computer Vision and
  Pattern Recognition ({CVPR})}}, 2019.

\bibitem{Zeng:etal:CORL2020}
Andy Zeng, Pete Florence, Jonathan Tompson, Stefan Welker, Jonathan Chien,
  Maria Attarian, Travis Armstrong, Ivan Krasin, Dan Duong, Vikas Sindhwani,
  et~al.
\newblock Transporter networks: Rearranging the visual world for robotic
  manipulation.
\newblock In {\em {Conference on Robot Learning ({CoRL})}}, 2020.

\bibitem{Zeng:etal:ICRA2018}
Andy Zeng, Shuran Song, Kuan-Ting Yu, Elliott Donlon, Francois~R Hogan, Maria
  Bauza, Daolin Ma, Orion Taylor, Melody Liu, Eudald Romo, et~al.
\newblock Robotic pick-and-place of novel objects in clutter with
  multi-affordance grasping and cross-domain image matching.
\newblock In {\em {Proceedings of the {IEEE} International Conference on
  Robotics and Automation ({ICRA})}}, 2018.

\bibitem{Zeng:etal:ICRA2017}
Andy Zeng, Kuan-Ting Yu, Shuran Song, Daniel Suo, Ed~Walker, Alberto Rodriguez,
  and Jianxiong Xiao.
\newblock Multi-view self-supervised deep learning for 6d pose estimation in
  the amazon picking challenge.
\newblock In {\em {Proceedings of the {IEEE} International Conference on
  Robotics and Automation ({ICRA})}}, 2017.

\end{thebibliography}

\end{document}